%% file: pprai_template.tex
\documentclass{pprai}

\usepackage[utf8]{inputenc}
\usepackage[T1]{fontenc}
\usepackage{graphicx}
\usepackage{amsthm}
\usepackage{txfonts}
\usepackage{url}
\usepackage{algpseudocode}
\usepackage[font=small,labelfont=bf]{caption}
\usepackage{wrapfig}
\usepackage[export]{adjustbox}

\title{Particle physics DL-simulation with control over generated data properties}
\headtitle{Particle physics DL-simulation with control over generated data properties}

\author{Karol Rogoziński$^{1}$, Jan Dubiński$^{1, 2}$, Przemysław Rokita$^{1}$, Kamil Deja$^{1, 2}$}
\headauthor{K.Rogoziński, J.Dubiński, P.Rokita, K.Deja}
\affiliation{%
  $^1$Warsaw University of Technology\\
  $^2$IDEAS NCBR\\
  rogo.karol@gmail.com}

\keywords{generative machine learning, high energy physics, ALICE, CERN}

\begin{document}
\maketitle

\begin{abstract}
The research of innovative methods aimed at reducing costs and shortening the time needed for simulation, going beyond conventional approaches based on Monte Carlo methods, has been sparked by the development of collision simulations at the Large Hadron Collider at CERN. Deep learning generative methods including VAE, GANs and diffusion models have been used for this purpose. Although they are much faster and simpler than standard approaches, they do not always keep high fidelity of the simulated data. This work aims to mitigate this issue, by providing an alternative solution to currently employed algorithms by introducing the mechanism of control over the generated data properties. To achieve this, we extend the recently introduced CorrVAE, which enables user-defined parameter manipulation of the generated output. We adapt the model to the problem of particle physics simulation. The proposed solution achieved promising results, demonstrating control over the parameters of the generated output and constituting an alternative for simulating the ZDC calorimeter in the ALICE experiment at CERN.
\end{abstract}

\section{Introduction}

The Large Hadron Collider is the most important instrument at CERN. It is the world's most powerful particle accelerator and is part of the CERN accelerator complex. Particle beams collide at four locations around the LHC, corresponding to the four main detectors: the LHCb~\cite{alves2008lhcb}, ATLAS \cite{atlas}, CMS \cite{cms} and ALICE \cite{alice}. In this work, we focus on the latter, and in particular its Zero Degree Neutron Calorimeter (ZDC) \cite{alice_report}, which collects the energy of observers, i.e., particles that did not participate directly in the collision. It is made of optical fibers arranged in a 44x44 grid.
From a physicist's perspective, simulations of calorimeter responses are just as crucial as experiments. If the results of simulations align with actual data, they can be considered as the real course of physical events and used for scientific research and analysis. Traditional methods of conducting simulations are costly and time-consuming \cite{Paganini:2017dwg}, so researchers have been developing alternative, more efficient methods for simulating collisions. This is further described in \cite{dubiński2023selectively, sinkhorn_deja}.

The objective of this work is to develop a machine learning model for rapid simulation of the ZDC calorimeter in the ALICE experiment, while also enabling control over the properties of the generated data. To achieve this, we utilized the CorrVAE model proposed in \cite{corrvae}. 
The method was initially designed to create images based on user-defined parameters, such as color, shape, or object position, and showed promising results. In this paper, we propose to adapt it to the high-energy physics (HEP) scenario. To that end, we introduce a new architecture that aligns with the ZDC response requirements, and introduce additional encoder with conditional latent space to allow conditional generations from particle properties.

In our experiments we show that our approach offers an alternative to current simulation methods, with a particular focus on controlling the response properties of particles, as presented in \cite{corrvae}. By assuming the physical parameters of the particles, the model is able to generate calorimeter responses that are both realistic and modifiable in a predetermined manner.

\section{Related Work}

To address the issues of the computational complexity of HEP simulations, researchers actively explore the possiblity of using generative machine learning methods instead. In \cite{metody_generatywne}, the authors simulated an electromagnetic calorimeter from the LHCb detector using the Wasserstein GAN \cite{wgan} model. To better represent momentum and position, the authors utilized a generator, discriminator, and an auxiliary regressor network to evaluate consistency. A similar idea was also presented in \cite{elektro_gan}, where the authors also used additional regressors for measuring the energy and position of the collisions.

\paragraph{The simulations of the Zero Degree Calorimeter}

Regarding the ZDC calorimeter of the ALICE experiment, in \cite{sinkhorn_deja}, authors introduced a method that based on Sinkhorn's autoencoder architecture \cite{sinkhorn}. It treats the calorimeter simulation problem as generating an image with the dimensions of a neutron calorimeter ZDC, which is 44x44. \cite{sinkhorn_deja} replaces the classical regularization in the latent space of VAE models with an additional noise generator defined as a neural network. The latest work on the subject is \cite{janek}, which presents two new approaches to solving the problem. The first method is based on Conditional VAE \cite{cvae}. The second method, based on Conditional GAN \cite{cgan}, also incorporates conditional information into the learning and generation process. Additionally, similar to the approach in \cite{metody_generatywne}, the authors utilized an auxiliary regressor to verify the accuracy of the particle's position. Another work that focuses on simulating the ZDC calorimeter \cite{dubiński2023selectively} employs a GAN network with added selective diversity increase loss regularization. 

\paragraph{Controlling properties in generative modeling}

Generating data with specific properties has been the subject of much work \cite{vae_1, graph_1, graph_2, corrvae, pcvae, speech}. The main concept behind those work is to combine variational autoencoders \cite{vae_1, pcvae} with various techniques such as graph networks \cite{graph_1, graph_2}. However, all the works mentioned have the problem of generating examples with given properties that are correlated with each other. This problem is solved by \cite{corrvae}, which served as the basis in the development of this work. 

\section{Problem Formulation}

The task is to generate simulation responses based on information about the particle at the time of the collision. To that end, we use a dataset described in \cite{hep}, which includes responses generated by classical Monte Carlo methods. The dataset contains nearly 300,000 simulated responses, each with 44x44 dimension, and includes nine properties for each particle: momentum and velocity in three planes, as well as mass, charge, and energy.

\section{Method}

The CorrVAE model was used as a starting point for this work. Its structure is based on the classical VAE, but it introduces two latent spaces instead of one. The first, $w$, is responsible for the image properties that the user has indicated, such as size and colour. The second latent space, $z$, is responsible for all other image features. These spaces are independent of each other, and the information they encode does not intersect in any way. Additionally, the authors proposed a mask with the same number of rows as the user-defined parameters and the same number of columns as the hidden space elements. A value of 1 in their intersection indicates that the element encodes a specific property. During the training phase, the property decoder receives the elements of the $w$ latent space multiplied by the mask. The decoder's output is then compared to the original user-defined properties using MSE as an additional loss function.

\paragraph{Basic adaptation for ZDC simulation}

To adapt the CorrVAE model for the simulation problem, several modifications were made. These changes were necessary due to the different specificity of the data compared to what was assumed in the original model. The modifications included changing the activation function in the last layer from Sigmoid to ReLU and replacing the reconstruction loss function from Binary Cross Entropy (BCE) to Mean Squared Error (MSE) due to the wider range of data generated by the simulation. The convolutional layers had to be adjusted to a size of 44x44, which is unusual for image generation problems.

\begin{figure}[t]
    \centering \includegraphics[width=0.60\linewidth]{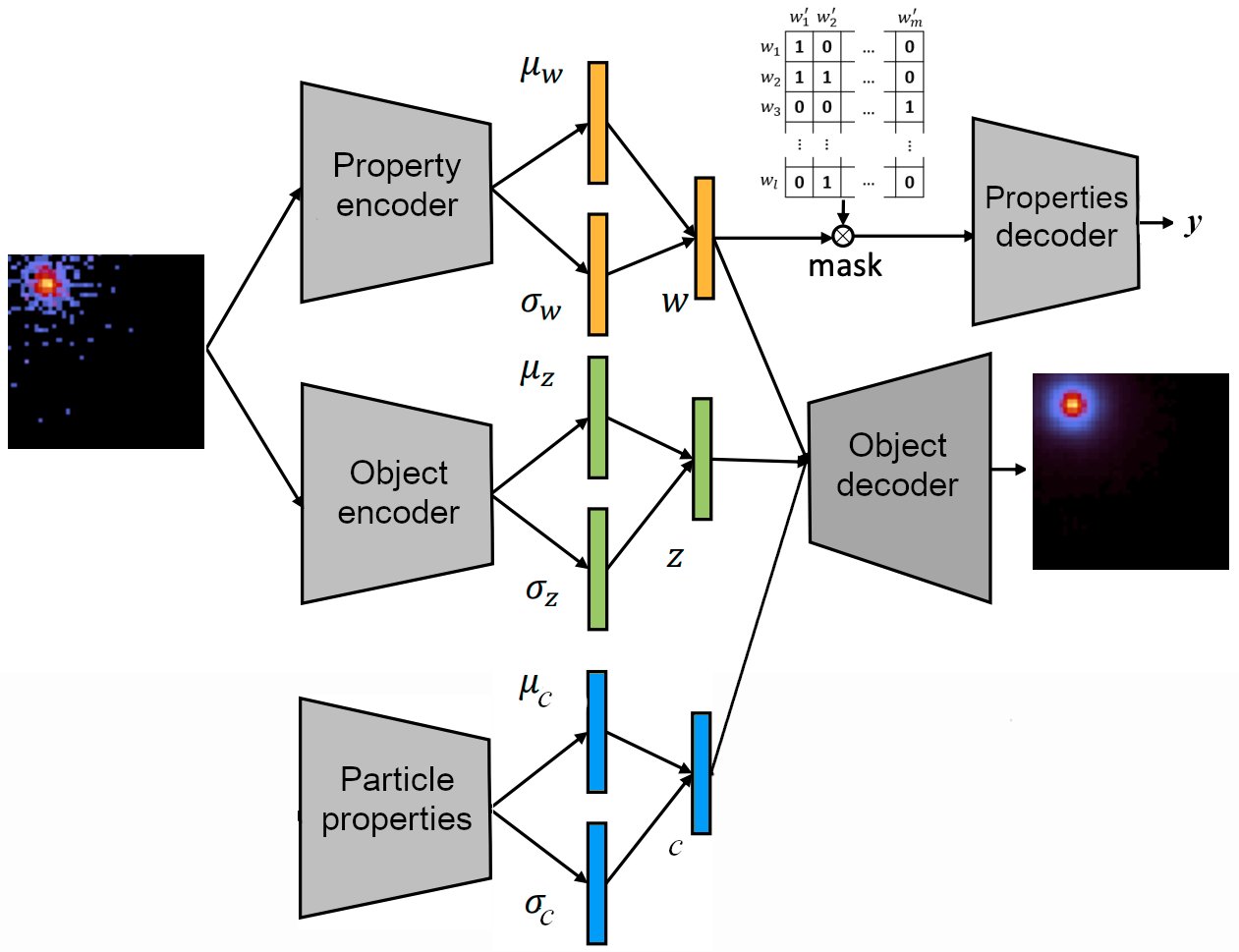}
    \caption{Model structure presentation. Unlike the classic CorrVAE, the information is encoded in three latent spaces, where the additional encoder encodes particle properties.}
    \label{figure:model}
\end{figure}

\paragraph{Model modifications}

Additionally, we modify the baseline model by adding conditioning to generate responses from the particle data in real-world conditions. However, a straight-forward conditioning is not compatible with the CorrVAE model, because the model assumes the independence of the hidden spaces and the ability to control the defined parameters with the hidden space $w$. Therefore, an additional encoder was added along with the hidden space $c$. The input of the encoder was the particle data, and the output stored information about the values and size of the response. The hidden space $w$ was only responsible for its position. New responses were generated by inputting data about the particle into the vector $c$ to encode its physical properties, optimizing the vector $w$ with a property decoder to include information about its position, and sampling the vector $z$ from a normal distribution. The final model structure is shown in Figure \ref{figure:model}.

\section{Experiments}
In this section, we provide the experimental evaluation of our method. Beside the standard evaluation of image reconstruction quality, the ZDC calorimeter simulation presents an unconventional case where we can directly measure the physical properties of the generated outputs. 
To that end, we follow~\cite{sinkhorn_deja}, and calculate values for five different channels that are used for physical analysis. We compare the distribution of those channels between generated and true images, with the Wasserstain distance, and average it across all channels. 

\begin{wrapfigure}{r}{0.58\linewidth}
\vskip  -8pt
    \centering \includegraphics[trim={0.5cm 0cm 0cm 0cm},clip,width=0.99\linewidth]{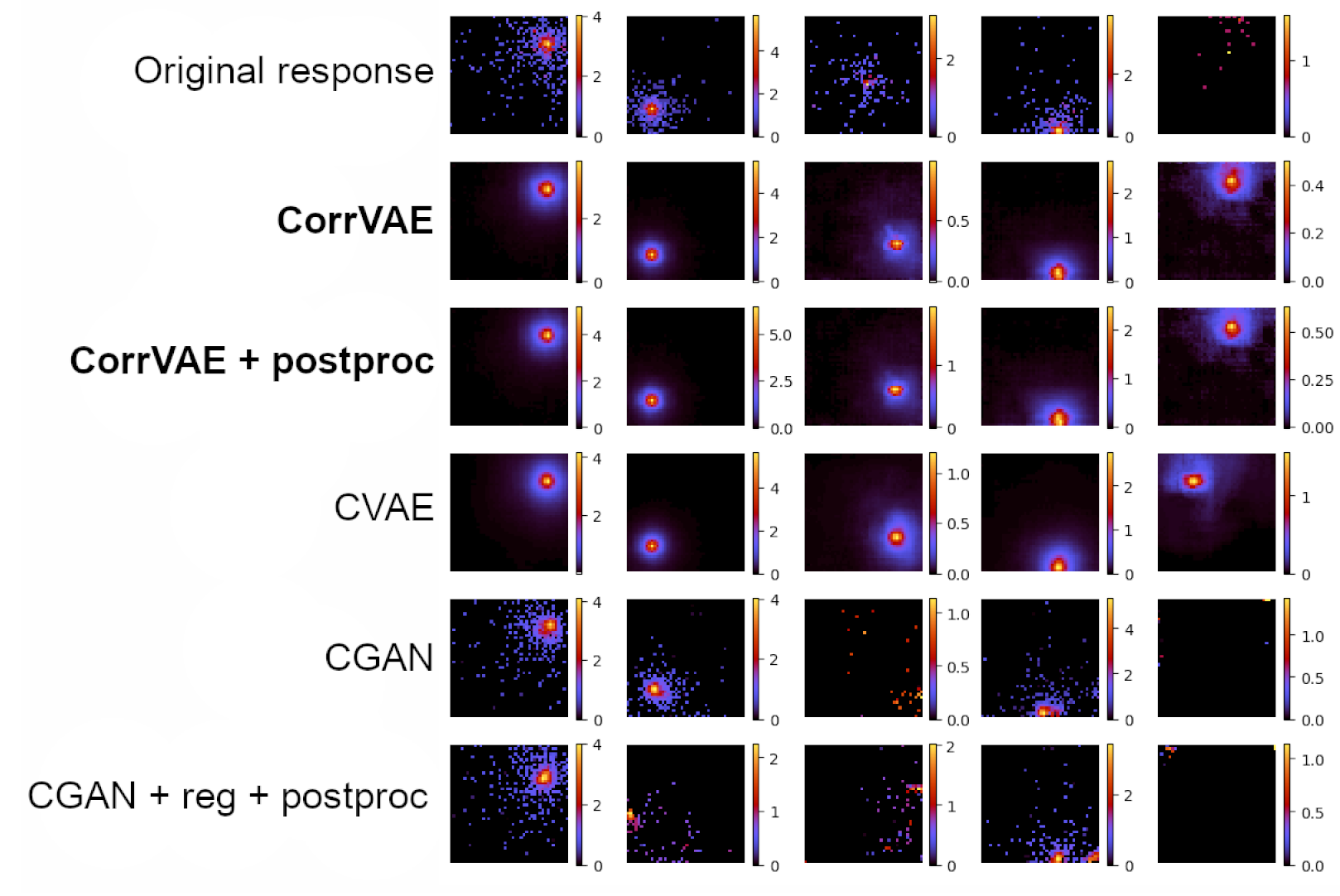}
    \caption{Comparison of randomly selected simulations generated by different models.}
    \label{figure:gen}
\vskip  -16pt
\end{wrapfigure}
We compare the results of different simulation approaches in Tab.~\ref{wyniki}. Additionally in Fig.~\ref{figure:gen} we show that our method yields much better alignment to the original responses (top row), when analysing the position of the center of the collision. However, similarly to other VAE-based models, we can see slightly blurred responses, which is the opposite of what happens in models based on the GAN architecture.

\begin{table}[t]
\small
\centering
\renewcommand{\arraystretch}{}
\begin{tabular}{p{5cm}cc}
\hline
 & MSE & Wasserstein \\
\hline
\textbf{CorrVAE} & \textbf{1.03} & \textbf{16.15} \\
\textbf{CorrVAE + postproc} & \textbf{1.18} & \textbf{3.83} \\

CVAE & 1.02 & 6.35 \\

CGAN & 2.96 & 8.27 \\

CGAN + reg + postproc & 2.98 & 5.15 \\
\hline
\end{tabular}
\caption{Comparison of results for the HEP dataset. Initially weaker than the other adapted models, the adapted CorrVAE model achievs the best results after applying post-processing. Values averaged over 3 runs.}
\label{wyniki}
\end{table}

\paragraph{Controlling the properties}



\begin{figure}[h]
    \centering \includegraphics[width=0.6\linewidth]{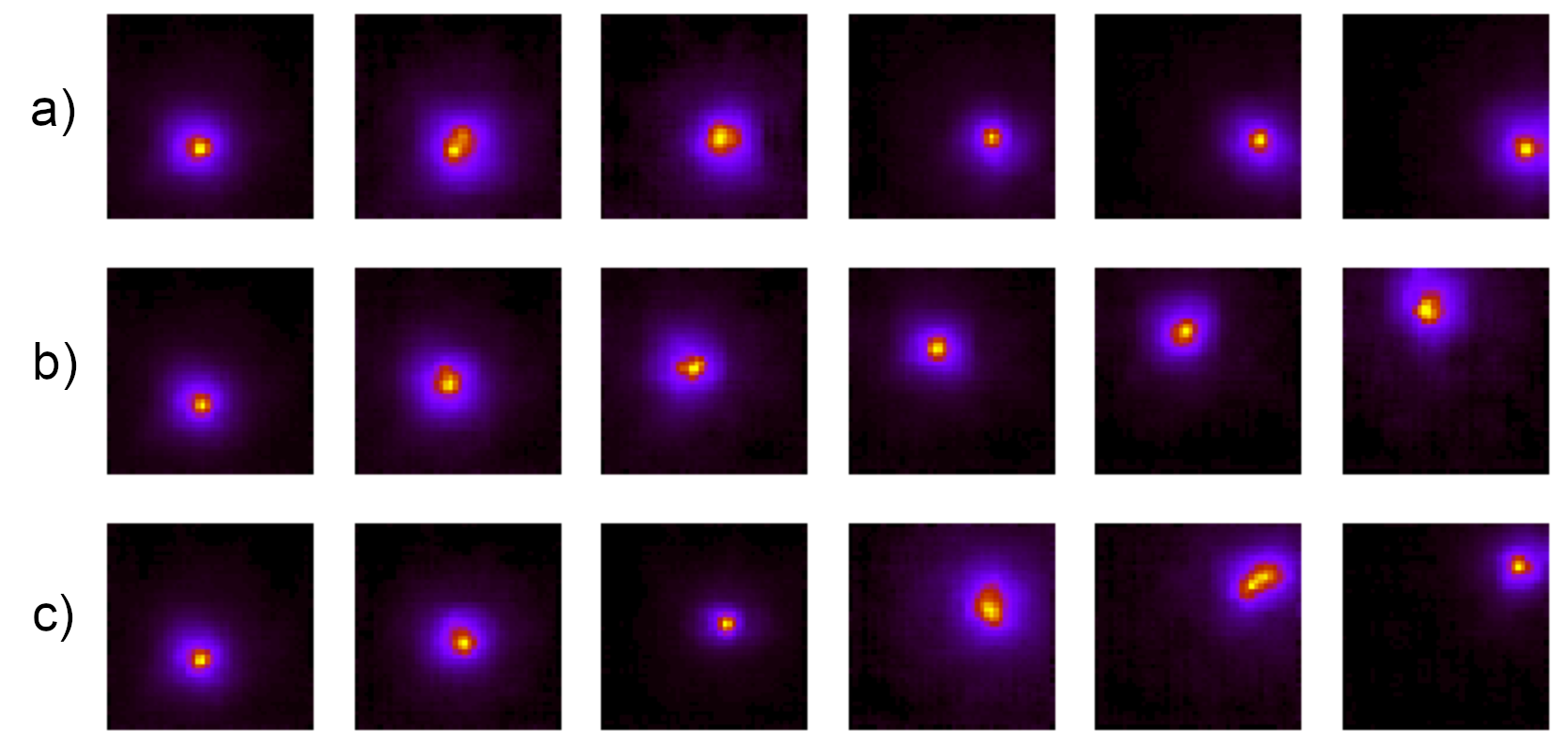}
    \caption{Generated images of presented model by traversing two latent variables in $w$ for HEP dataset according to the mask between x, y position and eight-element $w$ vector. (a) Traversing on the $w_1$ that controls x position; (b) Traversing on the $w_2$ that controls y position; (c) Traversing on the $w_1$ and $w_2$ at the same time.}
    \label{figure:przemieszczanie}
\end{figure}

To show the control over the properties of generated responses with our method, in the final experiment we show how we can adjust the position of the center of the mass, by interpolating between the selected values of the latent space. In particular, we train the model in a way that only selected two dimensions of the latent space $w$ are correlated with the position of the center of the mass ($w_1$ for $x$ position and $w_2$ for $y$ position). In Fig.~\ref{figure:przemieszczanie}, we show the effect of traversing through those dimensions.


\section{Conclusions}

In the paper, we successfully adapt and develop the CorrVAE model for high-energy physics simulation. This includes 
introduction of a new architecture, with additional conditional space needed to generate new simulations from the particle data. 
Our approach enables control over individual properties, which could be manipulated by feeding synthetic parameters or using the original particle properties at the time of the inference. The model prepared in this way can compete with the best methods currently in use.

\section*{Acknowledgments}
 
This research was funded by National Science Centre, Poland grants: 2020/39/ O/ST6/01478 and 2022/45/B/ST6/02817. This research was supported in part by PLGrid Infrastructure grants: PLG/2023/016393, PLG/2023/016361, PLG/2023/ 016278.

\small
\bibliography{pprai}
\bibliographystyle{pprai}

\end{document}